# Transitive Expert Error and Routing Problems in Complex AI Systems

Forest Mars <mars@mlops.nyc>
January 1, 2026

## Abstract

Validated domain expertise reliably enhances judgment within its boundaries but creates systematic vulnerabilities at domain borders: experts confronting problems that resemble but causally differ from their training reliably underperform novices facing identical tasks. We term this phenomenon Transitive Expert Error (TEE) and distinguish it from general overconfidence or Dunning-Kruger effects: TEE requires legitimate expertise as a precondition. The mechanisms that enable reliable judgment within domains become cognitive liabilities when structural similarity masks causal divergence.

We identify two core mechanisms. Structural similarity bias causes experts to overweight surface features (shared vocabulary, apparent patterns, formal structure) while missing fundamental differences in causal architecture. Authority persistence maintains professional confidence across competence boundaries through social reinforcement and metacognitive failures; experts experience no subjective uncertainty because their pattern recognition operates smoothly on familiar-seeming inputs. These mechanisms intensify under three boundary conditions: shared vocabulary masking divergent processes, social pressure for immediate judgment, and delayed feedback preventing recalibration.

These findings extend to artificial intelligence routing architectures: MoE systems, multi-model orchestration, tool-using agents, and RAG systems or arguably, any system routing inputs to specialized processors. Such systems exhibit routing-induced failures (wrong specialist selected due to surface similarity) and coverage-induced failures (no appropriate specialist exists). Both manifest through structural similarity in routing, authority persistence in specialist outputs, and delayed feedback in training. The result is a specific hallucination phenotype: confident, coherent, structurally plausible but causally incorrect outputs at domain boundaries.

Unlike human expert systems where these mechanisms are cognitive black boxes, AI routing architectures make them architecturally explicit and therefore addressable. We propose interventions at the router level (multi-expert activation with disagreement detection), specialist level (boundary-aware calibration), and training level (coverage gap detection). TEE in artificial systems has detectable signatures including routing patterns, confidence-accuracy dissociations, and domain-inappropriate content that enable monitoring and mitigation. The same mechanisms that are cognitive black boxes in human decision making become explicit routing decisions in AI systems, making boundary violations detectable and tractable as a new form of error correction.

# 1. Introduction

## 1.1 Established Accuracy of Expert Judgment Within Well-Defined Domains

Expert judgment within established domains demonstrates remarkable accuracy and consistency across diverse professional contexts (Shanteau 1992). Chess masters reliably identify optimal moves within milliseconds (Chase & Simon 1973), medical radiologists achieve diagnostic accuracy exceeding 90% within their specialty areas (Norman et al. 2007), and experienced auditors consistently detect financial irregularities with high precision (Bonner & Lewis 1990). This domain-specific reliability reflects the development of sophisticated pattern recognition capabilities and domain-appropriate reasoning heuristics through extended deliberate practice (Ericsson & Smith 1991).

The cognitive mechanisms underlying expert performance are well-characterized in the expertise literature (Chi et al. 1981). Experts develop hierarchically organized knowledge structures that enable rapid recognition of meaningful patterns within familiar problem spaces (Chase & Simon 1973). These structures, which differ in novices, integrate both surface features and deeper causal relationships specific to their domains (Chi et al. 1981) allowing experts to bypass effortful analytical reasoning in favor of intuitive pattern-based judgments that prove highly accurate within their areas of specialization. (Shanteau 1992). Cognitive science views experts as different from novices in nearly every aspect of cognitive functioning (Ibid).

## 1.2 Systematic Failure of Expert Reliability in Cross-Boundary Contexts

Despite high accuracy within domains, expert judgment reliability deteriorates systematically when applied across domain boundaries as calibration cannot keep pace with information recall (Alba & Hutchinson 2000). Tetlock's landmark 20 year study of political and economic forecasting revealed that expert predictions were only marginally more accurate than random chance, with more prominent experts, those with greater media visibility and professional recognition, performing worse than their less famous counterparts. The same factors that establish expert authority within domains may actively impair judgment when extended beyond appropriate boundaries.

Medical contexts provide similar evidence of cross-domain judgment failures. Specialty bias research demonstrates that when presented with identical clinical cases, surgical specialists recommend interventions at dramatically higher rates than internists recommend conservative management, independent of patient-specific factors that should determine optimal treatment (Sah et al. 2016). This pattern suggests that professional training shapes not merely knowledge but also the fundamental heuristics through which clinical problems are framed and evaluated, which may be domain-appropriate but inappropriately generalized.

Transfer literature documents similar patterns across learning contexts (Barnett & Ceci 2002). Skills and knowledge frequently fail to generalize beyond their original contexts even when surface similarities suggest applicability (Gick & Holyoak 1980). Transfer research focuses primarily on skill acquisition rather than expert judgment under novel input more generally (Mestre 2005), leaving the mechanisms underlying expert cross-domain failures underexplored.

## 1.3 Legitimate Expertise as Cognitive Liability

The phenomenon has been recognized since antiquity as ultracrepidarianism, 'speaking beyond one's competence', from the Latin *ultra crepidam* (beyond the sandal). However, these classical formulations provide descriptive terminology without mechanistic explanation. While ultracrepidarianism identifies that experts overstep boundaries, it does not explain why boundary recognition fails, why experts exhibit higher confidence than novices on identical cross-domain tasks, or what cognitive mechanisms produce systematic rather than random errors. We fill this gap with what we term transitive expert error (TEE): the systematic misapplication of valid domain-specific reasoning frameworks to structurally different domains, where the reasoning remains sound within its original context but invalidated due to unrecognized differences in causal architecture, feedback dynamics, or boundary conditions.

The cognitive mechanism involves surface versus structural similarity in analogical transfer. Research on analogical problem solving demonstrates that while structural features determine appropriate solutions, retrieval of source analogues occurs through summation of activation from shared surface features (Holyoak & Koh 1987). This creates a specific failure mode for experts: when inputs share surface features with their training domain but possess different causal structures, pattern recognition systems activate confidently based on familiar cues while underlying causal models remain inappropriate.

Domain expertise involves hierarchically organized knowledge structures integrating surface features with causal relationships specific to particular domains (Chi et al. 1981). These structures enable pattern-based judgments that prove highly accurate within areas of specialization (Shanteau 1992). However, the sophistication of such domain-specific frameworks creates vulnerability at boundaries: the same specialized processing enabling within-domain performance leads to systematic errors when boundaries are crossed unrecognized. The defining feature of transitive expert error is that legitimate expertise itself becomes the source of systematic error (Hinds 1999).

## 1.4 Distinction from Overconfidence and Dunning-Kruger Effects

Transitive expert error differs fundamentally from overconfidence bias, which involves systematic miscalibration of confidence relative to actual performance within a single domain (Dunning et al. 2004). While the Dunning-Kruger effect has been shown to be largely a statistical artifact of regression to the mean (Krueger & Mueller 2002) showing that individuals at all ability levels regress their self-assessments toward average performance, with low performers overestimating and high performers underestimating relative to objective standards. TEE represents a qualitatively different phenomenon that cannot be reduced to measurement artifacts: experts possess genuine competence and appropriate confidence calibration in their expertise (Shanteau 1992) while simultaneously displaying poor calibration in adjacent domains.

TEE makes a mechanistic prediction that cannot be explained by regression artifacts: experts with validated high performance in domain A should demonstrate systematically lower performance than domain-naive individuals in domain B when surface similarity masks causal divergence. This is directional and non-regressive: competence in A actively degrades performance in B. DK is a calibration problem; TEE is a boundary-crossing problem.

The failure to recognize when domain boundaries have been crossed and previously valid reasoning frameworks no longer apply contrasts sharply with both heuristic bias and Dunning-Kruger effects, and is critical for understanding both the mechanisms underlying TEE and the interventions required to mitigate it. The same pattern recognition capabilities, abductive reasoning processes, and professional confidence that enable high performance within domains actively impair judgment when inappropriately extended across domain boundaries.

## 2. Theoretical Framework

### 2.1 Domain Transfer and Negative Transfer

#### 2.1.1 Near Versus Far Transfer: The Barnett and Ceci Taxonomy

Barnett and Ceci (2002) provide a comprehensive taxonomy distinguishing near transfer, where knowledge and skills apply across highly similar contexts, from far transfer, where application occurs across contexts that differ substantially in surface features, temporal context, functional context, or modality. Near transfer succeeds relatively frequently because source and target domains share both surface features and underlying structural relationships (Singley & Anderson 1989). Far transfer, by contrast, frequently fails when surface dissimilarities obscure underlying structural correspondences or in the absence of such 'hints' (Gick & Holyoak 1980).

The expert judgment context would seem to involve a particularly problematic form of far transfer: domains that share vocabulary, conceptual frameworks, and professional contexts while differing fundamentally in causal mechanisms, feedback dynamics, or boundary conditions. This creates what Holyoak and Koh (1987) term 'surface similarity without structural alignment' which are contexts where experts perceive sufficient similarity to trigger domain-appropriate reasoning notwithstanding fundamental hidden differences invalidate that reasoning.

#### 2.1.2 Negative Transfer as Active Interference in Judgment

Negative transfer occurs when prior knowledge actively interferes with performance in new contexts rather than simply failing to transfer positively (Singley & Anderson 1989). Classic demonstrations show that participants trained on one problem-solving task perform worse on structurally similar but superficially different tasks than untrained controls, indicating that prior learning creates interference rather than facilitation (Reed et al. 1974).

In expert judgment, negative transfer operates via the activation of domain-inappropriate schemas and heuristics (Novick 1988). When experts encounter problems triggering in-domain pattern recognition developed and the new domain is structurally isomorphic, experts outperform novices (Ibid). When similarity is superficial not structural however, novices still approach unfamiliar problems cautiously and may avoid systematic errors, experts confidently apply familiar frameworks, making them particularly vulnerable to unrecognized negative transfer effects (Hinds, 1999). While Novick (1988) shows that experts *can* see structure, Hinds (1999) observes that their expertise often blinds them to the need to look for it in the first place.

#### 2.1.3 Limits of Existing Transfer Research for Expert Judgment Under Uncertainty

While the transfer literature provides crucial theoretical foundation, existing research focuses primarily on skill learning and problem-solving in well-structured domains with clear feedback

(Barnett & Ceci 2002). Expert judgment under uncertainty involves complexities not adequately addressed in standard transfer paradigms: judgments involve irreducible uncertainty rather than deterministic solutions (Tetlock 2005), feedback may be delayed or ambiguous (Hogarth & Karelaia 2007), and social and institutional pressures shape both judgment processes and confidence calibration (Tetlock 1985). These factors create conditions where standard transfer mechanisms interact with expert-specific cognitive processes to produce the systematic patterns characteristic of transitive expert error.

TEE addresses the inverse problem from transfer research. While transfer theory (Barnett & Ceci 2002; Gick & Holyoak 1980) examines why learners fail to apply knowledge across domains that appear dissimilar but share underlying structure, and the pedagogical challenge of helping novices recognize connections, TEE examines why experts over-apply knowledge across domains that appear similar but differ in causal architecture viz. the epistemic challenge of preventing confident misapplication when surface similarities are equivocal.

| Dimension | Transfer Research | TEE Framework |
|---|---|---|
| Focus | Whether/when skills generalize | Why confident misapplication occurs |
| Core phenomenon | Superficial dissimilarities obscure deep structural alignment | Superficial similarities mask deep structural divergence |
| Prediction | Transfer may fail | Failure exhibits specific signature |
| Confidence | Not addressed | Experts more confident than novices |
| Mechanism | Interference, surface/structural mismatch | Dual-process coordination trap, authority persistence |
| Social factors | Not central | Institutional reinforcement essential |
| AI application | General guidance | Specific architectural interventions |
| Failure signature | Unspecified | Coherent absurdity (high confidence + causal incorrectness) |

## 2.2 Dual-Process Betrayal: The CoordinationTrap

Transitive expert error represents a distinctive failure mode in dual-process theories of cognition (Kahneman, 2011): System 1 (intuitive, pattern-based) and System 2 (deliberative, analytical) each contribute to systematic error rather than one compensating for the other's limitations. Both systems operate correctly in domain, but this same coordination prevents boundary recognition.

### 2.2.1 System 1: Highly Tuned Pattern Matching Suppressing Analytical Skepticism

Within domains of expertise, System 1 processes become highly refined through extended deliberate practice, enabling rapid accurate pattern recognition (Chase & Simon, 1973). Chess masters identify optimal moves within milliseconds by recognizing board patterns as meaningful chunks (de Groot, 1965). Radiologists detect subtle abnormalities through pattern matching that bypasses conscious analysis (Norman et al. 2007). These intuitive processes generate strong subjective confidence before deliberative reasoning can engage (Kahneman, 2011).

The efficiency and historical success of these processes creates earned automaticity; experts have learned, correctly, to trust intuitions that consistently prove accurate (Shanteau, 1992). Within domains, following intuitive responses produces better outcomes than subjecting every judgment to extended deliberative analysis (Klein, 1998).

When domain boundaries are crossed without recognition, these same System 1 processes generate inappropriate pattern matches based on surface similarities (Holyoak & Koh, 1987). Cross-domain pattern matches produce the same phenomenological experience as valid within-domain recognition: rapid, confident, subjectively compelling intuitions. The expert experiences no qualitative difference between recognizing a genuine pattern and misrecognizing a superficial similarity across domains (Hinds, 1999).

This creates active suppression of analytical skepticism. The speed and confidence with which System 1 generates cross-domain matches preempts System 2 processes that might otherwise question domain applicability (Kahneman, 2011). Unlike novices, whose weak or ambiguous System 1 signals trigger deliberative analysis, experts experience strong intuitive responses that make deliberative checking feel unnecessary (Fleming & Lau, 2014).

### 2.2.2 System 2: Failure of Deliberative Checking Due to Home-Domain Confidence

System 2 might theoretically correct System 1's cross-domain pattern matching through explicit analysis of domain boundaries, causal mechanisms, and scope conditions (Kahneman 2011). However, System 2 engagement is metacognitively triggered by feelings of uncertainty, unfamiliarity, or complexity exceeding intuitive processing capacity (Fleming & Lau 2014).

In transitive expert error, these metacognitive triggers fail to fire. The expert's successful domain history creates calibrated confidence in System 1 processes (Shanteau 1992). When System 1 generates confident cross-domain matches, System 2 has no metacognitive reason to engage. The signals that would normally trigger deliberative checking are absent because System 1 is generating outputs indistinguishable from valid within-domain judgments (Hinds, 1999).

The failure is not a defect in System 2's capacity but in its deployment. Within domains, selective deployment is adaptive; experts correctly conserve cognitive resources by relying on well-calibrated intuitions (Klein 1998). Across domains this same selective deployment becomes maladaptive as metacognitive signals are suppressed by confident System 1 pattern matching.

### 2.2.3 The Coordination Trap: Mutual Reinforcement of Domain Misalignment

Both systems operate exactly as they should within domains, but their coordination prevents boundary recognition. System 1 generates confident pattern matches, System 2 trusts these matches because they feel identical to valid intuitions, and the absence of uncertainty signals prevents the deliberative analysis that might recognize domain mismatch.

This coordination trap distinguishes TEE from standard dual-process failures. In typical heuristic-driven biases, System 1 generates quick responses that System 2 should but fails to override (Kahneman, 2011). The intervention is to strengthen System 2 engagement to slow down, think carefully, check intuitions. In transitive expert error, the problem is more fundamental: System 2 has no reason to engage because System 1 provides exactly the kind of reliable-seeming output that System 2 has learned to trust through years of domain experience.

The expert cannot simply think more carefully because careful thinking itself operates within domain-specific frameworks (Chi et al. 1981). When an expert deliberatively analyzes a cross-domain problem, they deploy analytical tools and causal models from their home domain: the very frameworks that structural similarity has inappropriately activated. System 2 engagement does not correct the error; it elaborates and justifies it through explicit reasoning trapped within the wrong domain model.

This explains why transitive expert error survives sustained reflection and explicit justification. The expert is not being careless or failing to think critically. Both intuitive and deliberative processes operate at full capacity, but both operate within an inappropriately activated domain framework. Neither process has access to metacognitive signals indicating domain mismatch.

## 2.3 Mechanisms of Transitive Expert Error

### 2.3.1 Structural Similarity Bias and Abductive Interference

**Overweighting Surface Similarities Relative to Causal Architecture**
Structural similarity bias involves systematic overweighting of surface-level similarities while underweighting or ignoring fundamental differences in causal architecture between domains (Holyoak & Koh 1987). Experts develop sophisticated capabilities for identifying meaningful patterns within their domains, but these capabilities rely on learned associations between surface features and underlying causal relationships specific to those domains (Chi et al. 1981).

When experts encounter problems in adjacent domains that share surface features (eg. similar vocabulary, analogous conceptual frameworks, or parallel professional contexts) their pattern recognition systems activate domain-specific schemas based on these surface similarities (Gentner & Markman 1997). However, if the causal architecture underlying these surface features differs fundamentally between domains, schema activation generates inappropriate inferences and predictions.

**Abductive Inference as Liability Through Forced Explanatory Fit**
Expert reasoning within domains relies heavily on abductive inference: reasoning to the best explanation given observed patterns (Peirce 1931). Through extended experience, experts develop rich repertoires of explanatory models that connect observable patterns to underlying causal mechanisms within their domains (Chi et al. 1981). When confronted with ambiguous or incomplete information, experts reflexively generate explanatory hypotheses that best fit the observed data given their domain models.

In cross-domain contexts, this abductive reasoning process becomes a liability. Rather than recognizing unfamiliarity and withholding judgment, experts actively "force" explanatory fits by selecting the best-matching model from their domain-specific repertoire (Hinds 1999). This forced fit generates confident explanations that may be internally coherent within the expert's framework but fundamentally misaligned with the actual causal structure of the target domain. The abductive process operates at full strength, providing subjectively compelling explanations that preempt recognition of domain mismatch.

**Professional Identity as Pattern-Finding Compulsion**
Expert professional identity is fundamentally organized around pattern recognition and explanation (Shanteau 1992). Experts are trained, socialized, and rewarded for their ability to identify meaningful patterns where others see only noise and to generate compelling explanations for complex phenomena. This professional identity creates what might be termed a 'pattern-finding compulsion,' a strong professional and psychological pressure to find patterns and generate explanations rather than acknowledge unfamiliarity or withhold judgment.
When confronted with cross-domain problems, this pattern-finding compulsion drives experts to apply their pattern recognition capabilities even when appropriate caution would suggest restraint. The alternative admitting inability to recognize patterns or generate explanations, conflicts with core professional identity and social role expectations (Tetlock 1985). This creates systematic pressure toward overextension of domain-specific reasoning rather than appropriate humility at domain boundaries.

### 2.3.2 Authority Persistence and Metacognitive Myopia

**Social and Institutional Reinforcement of Authority Across Domains**
Expert authority operates through multiple social and institutional mechanisms that extend beyond individual cognitive processes (Tetlock 2005). Professional credentials, institutional affiliations, and past successes create generalized authority that audiences attribute to experts across contexts (Shanteau 1992). This social authority creates expectations that experts can provide authoritative judgments not only within their narrow specialties but across related areas where their expertise might plausibly apply.

These social dynamics create systematic pressures for experts to extend their reasoning across domain boundaries rather than acknowledging limitations (Tetlock 1985). Institutional contexts such as legal proceedings, policy consultations or media commentary often demand immediate expert judgment without time for appropriate boundary recognition or qualification. The expert who acknowledges uncertainty or unfamiliarity faces professional costs, while confident cross-domain judgments receive social reinforcement even when they prove inaccurate (Tetlock 2005).

**Confidence Without Calibration and the Metacognitive Halo Effect**
The metacognitive processes that enable appropriate confidence calibration within domains fail to operate effectively across domain boundaries (Fleming & Lau 2014). Within domains, experts develop accurate metacognitive monitoring through extensive feedback that correlates confidence with accuracy (Shanteau 1992). This creates justified confidence in domain-appropriate judgments and appropriate uncertainty when domain-relevant cues are absent or ambiguous.

However, when domain boundaries are crossed without recognition, experts experience what might be termed a 'metacognitive halo effect': the justified confidence associated with domain expertise extends to judgments about structurally different problems that share surface similarities. The metacognitive signals that would normally indicate knowledge limitations, but such feelings of uncertainty, recognition of missing information, or awareness of reasoning difficulties are attenuated or absent because the surface similarities trigger pattern recognition responses (Hinds 1999).

### 2.3.3 Boundary Conditions: The Perfect Storm

Transitive expert error occurs most frequently and severely when three conditions coincide:

**Shared Vocabulary Masking Divergent Processes**
Domains that share vocabulary while differing fundamentally in causal mechanisms create what linguists call *false friends*: words that appear identical but carry totally different meanings (Chamizo 2008). Terms like 'risk,' 'transmission,' 'network,' and 'optimization' trigger domain-specific pattern recognition while underlying referents diverge. An infectious disease expert's 'transmission' incorporates biological pathways and population mixing; a financial analyst's 'transmission' incorporates information flow and systemic contagion. The shared terminology obscures fundamental differences, enabling confident application of inappropriate models.

**Social Pressure for Immediate Judgment**

Institutional contexts often demand immediate expert judgment without time for boundary recognition or qualification (Tetlock 1985). Crisis contexts (health epidemics, financial instability, security threats) create strong demands for expert guidance despite inherent uncertainty. The expert who acknowledges knowledge limits faces professional costs, while confident cross-domain judgments receive institutional reinforcement even when based on inappropriate domain extensions (Tetlock 2005). These pressures operate against metacognitive processes that enable boundary recognition (Fleming & Lau 2014).

**Delayed or Ambiguous Feedback**

Applications often occur in "wicked" contexts where feedback on judgment accuracy is delayed, ambiguous, or absent entirely (Hogarth & Karelaia 2007). Policy recommendations produce effects over years; complex systems involve multiple causal factors operating simultaneously, obscuring the contribution of any single judgment. Absence of clear feedback prevents the learning that might enable experts to recognize competence boundaries and adjust reasoning accordingly (Tetlock 2005). Within domains, expert learning relies on relatively immediate feedback that enables calibration; across domains, this feedback mechanism fails.

# 3. Implications and Interventions

Institutional attempts to formally define expertise, such as the Daubert standard, don't prevent TEE because they prioritize methodology validation while ignoring the bounded rationality (Simon, 1947) that forces even high-level experts to rely on domain-specific 'chunks' and heuristics (Chase & Simon, 1973). This creates an authority gap where legal and policy systems treat expertise as a portable credential rather than a boundary-contingent capability. Daubert assumes that a reliable method ensures a reliable conclusion; however, as Simon established decades prior, the expert's cognitive limits ensure they will confidently project that 'reliable' method onto structurally foreign domains wherever surface patterns appear to justify it.

Academic peer review, corporate consulting, and policy advisory boards all rely on expert authority without systematic mechanisms to detect boundary violations. A neuroscientist's expertise in synaptic plasticity may lead to confident but causally inappropriate claims about organizational learning. An economist's mastery of equilibrium models may produce confidently wrong predictions about far-from-equilibrium transitions. Mitigating this requires epistemic auditors trained to detect lexical ambiguity, shared vocabulary masking divergent causal architectures, ensuring the expert's cognitive shortcuts remain within their validated operational scope (Simon, 1962). These auditors don't evaluate the expert's domain competence but instead verify that boundary conditions hold, analogous to statistical consultants who verify appropriate use of methods rather than domain content, but focused on boundary detection rather than methodological correctness.

However, interventions targeting human expert systems face fundamental tractability constraints. The mechanisms producing TEE, System 1 pattern matching operating below conscious awareness, System 2 failing to engage due to absent metacognitive triggers, professional identity pressures reinforcing pattern-finding compulsion, resist direct modification. Experts cannot simply "recognize boundaries" when the dual-process coordination trap prevents the very signals that would trigger boundary recognition. Institutional scaffolding such as epistemic auditors provides external forcing functions, but the underlying cognitive mechanisms remain opaque black boxes observable only through behavioral outcomes.

This tractability problem motivates extending the TEE framework to artificial expert systems (Part II). Where human cognitive mechanisms resist intervention due to opacity, AI routing architectures make identical mechanisms architecturally explicit. Each intervention targets specific TEE mechanisms: multi-expert activation addresses structural similarity bias by forcing comparison between alternative domain models, boundary-aware calibration addresses authority persistence by training specialists to recognize competence limits, and meta-experts address coverage gaps by providing architectural capacity absent from domain specialists. The interventions are complementary: routing improvements reduce misrouting, calibration reduces overconfidence given correct routing, and meta-experts catch cases where neither applies. What remains intractable in human expert systems becomes addressable through architectural design in AI systems.

## 4. Case Study: Deterministic Models Applied to Reflexive Markets

The 2008 financial crisis illustrates transitive expert error operating at scale. Quantitative analysts with PhDs in physics, mathematics, and engineering developed sophisticated models for pricing derivatives and managing risk based on reasoning frameworks proven successful in their original domains (Derman 2004; Taleb 2007). However, financial markets are fundamentally reflexive systems where participant beliefs shape the phenomena being modeled, a canonical example of the double hermeneutic (Ferraro et al. 2005; Giddens 1976). The Black-Scholes-Merton model did not merely describe option pricing; it created a pricing regime where market behavior conformed to the model because traders used it (MacKenzie 2006). Models that successfully predicted physical system behavior failed catastrophically when applied to markets because surface similarities masked fundamental differences in causal architecture.

### 4.1 Surface Similarities That Triggered Pattern Recognition

The structural parallels were substantial. Both domains employed stochastic differential equations, probability distributions, and optimization frameworks. The Black-Scholes-Merton option pricing model used heat equation mathematics directly from physics (Black & Scholes 1973). Value-at-Risk models employed statistical mechanics approaches (Jorion 2006). Financial time series exhibited statistical regularities (volatility clustering, fat-tailed distributions) comparable to physical phenomena. The vocabulary reinforced apparent similarity: 'equilibrium,' 'stability,' 'volatility,' 'momentum,' and 'energy' appeared in both contexts. The mathematical structures genuinely matched.

### 4.2. Causal Divergences That Invalidated the Transfer

*Fundamental causal architectures diverged in ways that invalidated the reasoning frameworks:*

**Reflexivity**: Physical systems exhibit external causation; particles don't change behavior based on physicists' models. Financial markets are reflexive: participant beliefs directly influence the phenomena being modeled (Soros 1987; MacKenzie 2006).

**Non-Ergodicity**: Physical systems are typically ergodic: time averages equal ensemble averages. Financial markets are non-ergodic: a single trajectory can be ruined even when ensemble statistics appear favorable (Peters & Gell-Mann 2016). The crisis manifested non-ergodic dynamics where extreme events are structurally intrinsic, not anomalous.

**Endogenous Risk**: In physical systems, risk comes from external perturbations. In financial systems, risk emerges from collective participant behavior (Danielsson 2002). When many participants followed similar risk models and simultaneously reduced exposure, they generated the market crashes their models predicted as improbable (Bookstaber 2007).

**Regime Changes**: Physical systems exhibit stationary statistical properties. Financial systems undergo regime changes where statistical properties shift discontinuously (Taleb 2007). Models calibrated on stable periods systematically underestimated risk during transitions because they assumed stationarity intrinsic to physics but absent from reflexive markets.

## 4.3 TEE Mechanisms in Operation

Surface similarities triggered confident application of physics-derived frameworks because the mathematical structures genuinely matched. When confronted with financial time-series data, quants generated explanations using sophisticated tools, stochastic processes, equilibrium models that fit historical data beautifully. The abductive process selected models from the physics repertoire because those models had proven powerful in structurally similar-appearing contexts.

Authority persistence operated through institutional reinforcement. During 1990s–2007, quants received massive validation via high salaries, prestige, and apparent market success (Patterson 2010). Model failures were interpreted as parameter estimation issues, not fundamental domain mismatch. All three boundary conditions were present: shared vocabulary masking divergent processes, crisis pressure creating institutional demand for confident quantitative guidance, and delayed feedback. Catastrophic failures only appeared in 2008, years after model adoption.

This case illustrates TEE's defining features. The quants possessed genuine expertise; within physics, their frameworks were valid. The error emerged from extending legitimate expertise across a boundary where surface similarities masked causal divergences. Novices without physics training would not have built these sophisticated models that failed. The expertise itself became the source of systematic error.

# Part II: Implications for Routing Architectures in AI Systems

## 5. From Human to Machine Expert Systems

The mechanisms underlying Transitive Expert Error in human judgment map directly onto mixture-of-experts (MoE) transformer failures. When a physicist confidently applies deterministic models to reflexive financial markets, the error stems from pattern recognition optimized for one causal structure misfiring on superficially similar but fundamentally different phenomena. MoE systems exhibit the same pattern: specialized modules confidently execute domain-appropriate reasoning on inputs where surface similarities mask causal divergence.

MoE architectures consist of specialized subnetworks and a gating mechanism that directs inputs to appropriate experts (Jacobs et al. 1991; Shazeer et al. 2017). This explicitly instantiates the human expert system problem: given modules with bounded competence, how do you route problems to the right specialist? Both fail when surface features drive selection while causal differences go undetected. Modern architectures optimize for computational efficiency and load balancing but do not address domain boundary detection.

Any AI system that routes inputs to specialized processors faces the TEE problem: given modules with bounded competence, how do you select the appropriate specialist when surface features can mask causal divergence? The answer to this question is implemented across disparate modern AI systems:

**Mixture-of-experts**: Router gates direct tokens to expert subnets based on learned similarity

**Multi-model systems**: Orchestration layers select between specialized models (coding, analysis, creative tasks)

**Tool-using agents**: Selection mechanisms choose appropriate tools from available APIs

**Compound AI systems**: Meta-controllers route queries to specialized components (retrieval, generation, verification)

**RAG with specialized sources**: Query routing selects knowledge bases or document collections

Failure modes occur when surface similarity triggers wrong specialist selection or no appropriate specialist exists. The mechanisms are identical: structural similarity bias in routing decisions, authority persistence in specialist outputs, and delayed feedback preventing recalibration.

MoE architectures instantiate expert system architecture: specialized modules ("experts") with bounded competence and explicit routing mechanisms determining which expert processes which input. The framework and interventions we develop apply to any routing architecture; MoE systems are the purest case where the expert-routing structure is architecturally explicit. The failure examples presented are theoretical illustrations of how surface similarity could trigger TEE in MoE systems. While these specific cases require empirical validation in deployed systems, the underlying mechanism (routing based on token similarity activating domain-inappropriate experts) follows directly from MoE architectural design.

## 6. Architectural Mapping of TEE Mechanisms

### 6.1  Two Pathways to Failure

TEE explains a specific failure mode, not all MoE failures. It does not explain random errors from insufficient training, distributional shift within a single expert's domain, adversarial routing exploits, or corrupted weights. MoE systems exhibit TEE through two distinct architectural pathways, each with different mechanisms requiring different interventions. Both routing-induced and coverage-induced failures demonstrate how structural similarity bias, authority persistence, and delayed feedback operate at multiple architectural levels, reinforcing each other to produce systematic hallucination patterns at domain boundaries

- **Routing-induced TEE**: The wrong expert is selected. An appropriate expert exists, but surface similarity causes misrouting. The activated expert executes confidently within its own domain model, unaware the input requires different causal reasoning. Signature: single expert activation, high confidence, domain-inappropriate content.

- **Coverage-induced TEE:** No appropriate expert exists. The input falls into a gap between available specializations. The router must activate something due to architectural constraints (gating functions sum to one), selecting the least-dissimilar option. Signature: low similarity scores across all experts, high inter-expert disagreement, or forced consensus around an inappropriate model.

Understanding this distinction is critical for intervention design: routing failures require improved boundary detection and multi-expert disagreement signals; coverage failures require explicit abstention mechanisms and meta-expert arbitration..

### 6.1.1 Routing-Induced Failures

In routing-induced failures, structural similarity bias is implemented as a direct consequence of its objective function. Contemporary transformer-based mixture-of-experts routers assign each input to a subset of experts via a gating network producing sparse activation patterns (Shazeer et al. 2017; Fedus et al. 2022). Routing decisions are shaped by auxiliary objectives, such as load-balancing losses, that encourage experts to specialize to regions of the training distribution (Riquelme et al. 2021). At domain boundaries, this optimization produces systematic misrouting.

The failure emerges from geometric properties of the embedding space. Transformer embeddings exhibit anisotropy, with representations concentrated in narrow cones, reducing effective angular separation between semantically distinct inputs (Ethayarajh, 2019; Gao et al. 2019). Consider the query: 'How does the net handle sudden increases in load?' Tokens such as net, load, and handle occupy regions of latent space where statistical proximity reflects co-occurrence patterns rather than causal relations. When routing relies on similarity-based gating, inputs dominated by technical vocabulary may be assigned to a computer science expert based on superficial lexical overlap rather than domain appropriateness, consistent with shortcut-learning behavior (Allingham et al. 2023).

Once activated, the expert produces confident pattern completion within its learned distribution. Rather than generating nonsensical output, the model performs high-fidelity continuation optimized for likelihood under its internal statistics (Holtzman et al. 2021). This results in systematic interpretation of load as network throughput and net as graph topology, even when the broader context is misaligned, reflecting reliance on surface-form cues over structural constraints (McCoy et al. 2023). The output appears coherent while remaining causally misapplied.

This constitutes functional entrapment: the expert module operates as designed on familiar input patterns but lacks explicit mechanisms for signaling when applied outside its domain of competence (D'Amour et al. 2020). Because generation is optimized for low-perplexity continuation rather than domain validation, pattern-matching proceeds smoothly even under distributional mismatch, yielding high-confidence outputs despite invalid domain transfer (Kuhn et al. 2023).

### 6.1.2 Coverage-Induced Failures

Coverage-induced failures occur when inputs reside in regions of embedding space where no expert has specialized training. The router faces what might be termed an abstention gap: it must select experts even when no appropriate specialist exists. In continuous high-dimensional representations, 'out of distribution' is not binary but a distance metric. The sparsity constraint forces the router to assign inputs to the top-k experts with highest similarity scores, even when all similarities are low or when multiple experts show comparable but inappropriate matches.

This produces a nearest-neighbor trap. The router is mathematically constrained to activate some expert because gating functions must sum to one (or k in top-k routing). When the true answer is 'no available expert is appropriate,' the architecture provides no mechanism for this response. The system selects the least-dissimilar expert and proceeds with confident generation.

Consider queries about economic network effects, such as how market platforms create value through user base growth. This domain shares vocabulary with computer networking ('nodes,' 'connections,' 'network effects,' 'scaling') and mathematical graph theory ('topology,' 'centrality,' 'clustering'). A router might activate the networking expert, which applies protocol design principles and bandwidth optimization to market dynamics. Alternatively, it might activate the graph theory expert, which applies mathematical properties without accounting for strategic behavior or reflexive feedback in economic systems.

The case of Metcalfe's Law provides a historical parallel to MoE routing failures. Originally valid for telecommunications switching where each connection has roughly equal utility, the formula was aggressively transferred to internet business valuation during the dot-com era. This historically crashed out because economic networks exhibit divergent causal dynamics: not all connections create equal value, network effects show diminishing returns, and human reflexive behavior introduces feedback loops absent in technical systems (Briscoe, Odlyzko, & Tilly, 2006).

In both human and MoE contexts, the error mechanism is identical: functional entrapment. A technically sophisticated model is applied to a domain where underlying mechanisms diverge fundamentally, yet surface similarity triggers high-confidence application. As Odlyzko observed, the failure was not in the arithmetic of the quadratic scaling, but in the inappropriate domain transfer from network theory to economic contexts. Metcalfe eventually revised his formula toward a logarithmic scaling but only after the structural similarity between technical and economic nodes had contributed to systematic overvaluation and the subsequent market crash.

The activated expert produces outputs with high structural plausibility. Perplexity remains low because pattern completion operates on familiar token sequences. The technical sophistication of the output masks total divergence from the domain's causal reality. Current MoE architectures lack mechanisms for entropy-based abstention or confidence calibration to boundary distance. The expert must generate something, and it generates confidently because its internal processes operate smoothly even when domain application is inappropriate.

These mechanisms reinforce each other across the system. Routing errors based on structural similarity trigger expert activation. Experts generate confident outputs due to authority persistence. RLHF training fails to penalize these outputs because evaluators cannot detect the domain mismatch, preventing any corrective signal. The system has no architectural pathway for recognizing or learning from boundary violations.

## 6.2 TEE as a Hallucination Phenotype

"Hallucination" is currently used to describe a heterogeneous set of failure modes, including memorized training artifacts, token-level prediction errors, logical inconsistencies, and factual inaccuracies. Despite extensive usage, the term has historically lacked a unified theoretical framework capable of distinguishing failures that are merely incorrect from those that are systematically misleading.

Recent work by Liu et al. (2025) argues that existing definitions of hallucination such as 'content unsupported by source' (Ji et al. 2023), capture only restricted projections of a general failure mode. Much of this conceptual confusion arises from the proliferation of task-specific, partially incompatible definitions. By reviewing formulations across machine translation, summarization, and question-answering, they demonstrate that existing definitions, such as 'content unsupported by source' or 'content inconsistent with world knowledge,' capture only restricted projections of a more general failure mode. Liu et al. subsume these under a unified account, defining hallucination as an observable mismatch between a model's internal world representation and a designated reference world model.

Applying this framework, we propose that Transitive Expert Error (TEE) constitutes a distinct and particularly hazardous hallucination phenotype. Within the 'World Model' taxonomy, TEE represents a mismatch where the internal world model is locally coherent but globally misaligned with the target domain. We define this phenotype by the simultaneous presence of four characteristics:

**High confidence:** Low perplexity outputs, peaked probability distributions, and definitive, unhedged language.

**Structural plausibility:** Correct formal structure, appropriate technical vocabulary, internally coherent reasoning, and expert-like presentation.

**Causal incorrectness:** Misapplied models, inappropriate domain transfer, and false mechanistic explanations.

**Boundary localization:** Errors concentrate at interfaces between adjacent domains rather than appearing as random or diffuse inaccuracies.

The conjunction of these properties makes TEE-type hallucinations maximally dangerous. Unlike conventional hallucinations, which often exhibit detectable surface anomalies or epistemic uncertainty, TEE outputs preserve every outward marker of reliable expert judgment, confidence, fluency, and technical sophistication, while being fundamentally wrong at the level of causal structure.

This explains both the persistence of TEE under scaling and its resistance to standard hallucination detection methods. Most current mitigations target surface irregularities or factual inconsistency; however, because TEE is a boundary-level causal misalignment, it remains invisible to detectors that do not explicitly account for domain-specific causal architectures.

## 6.3 Why Standard Detection Methods Fail

*Standard detection methods fail not only at the surface level, but across the full stack of safety mechanisms, because TEE's characteristics subvert both shallow and deep detection approaches.*

**Perplexity-based uncertainty** quantification fails because TEE produces low perplexity by design. Tokens are lexically familiar and the sequence matches the expert's learned domain-specific patterns. As Kuhn et al. (2023) demonstrate, token-level entropy (and thus perplexity) measures predictive fluency rather than semantic truth. The expert executes confidently in the wrong domain because the input triggers high-probability shortcuts identified in its training data, a phenomenon where neural nets may assign high confidence to unrecognizable images (Nguyen et al. 2015) and LLMs may exhibit high internal consistency while remaining globally misaligned with the target domain (Grosse et al. 2023).

**Ensemble methods** fail because TEE errors are systematic rather than independent. When multiple experts are misrouted to the same inappropriate domain due to shared lexical triggers, voting produces a confident consensus around a fundamentally flawed answer. This violates the foundational requirement for ensemble efficacy: that individual members must exhibit diverse failure modes (Kuncheva, 2014). In MoE architectures, this diversity is often undermined by shared architectural priors and overlapping training distributions, leading to redundant failure rather than robust error correction (Zhuang et al. 2024). Consequently, the majority vote reflects a collective 'False Friend' hallucination rather than a truthful consensus.

**Embedding-distance** methods fail because they measure proximity rather than domain validity. Distance to training centroids captures statistical familiarity in representation space but cannot distinguish correct domain membership from semantic similarity to an inappropriate domain. Boundary cases remain close to the wrong domain manifold, which is precisely what produces TEE routing errors, making distance-based novelty detection unreliable (Shafaei et al. 2019). This failure is most acute for near-OOD inputs, which occupy dense regions of the embedding space and are therefore assigned high confidence instead of being flagged as anomalous (Winkens et al. 2020). Because LLM embeddings privilege lexical and semantic overlap over structural or causal constraints, inputs are preferentially mapped to the nearest available centroid even when the underlying domain model is wrong, yielding confident misclassification rather than uncertainty (Kirsch et al. 2022).

**Coherence metrics** fail because TEE outputs exhibit logical flow, appropriate transitions, and consistent terminology. Coherence measures structural quality, not causal correctness. This aligns with what MacArthur (2025) terms the 'fluency fallacy' and L. Zhang et al. (2025) call the 'coherence trap': grammatical polish and semantic coherence mask fundamental inaccuracies.

**Format validation** fails because TEE outputs satisfy structural expectations perfectly. LLMs exhibit a well-documented surface-form bias, prioritizing the completion of high-probability templates and formal patterns even when the semantic content instantiated within those forms is causally disconnected from the prompt (McCoy et al. 2023). Because the MoE router has activated a domain specialist, the resulting output utilizes the precise formatting, syntax, and stylistic markers of that domain (e.g., LaTeX formulas, specialized headers). Adherence to such canonical forms systematically masks semantic error, allowing structurally correct but domain-misapplied outputs to pass automated format checks that implicitly conflate formal validity with technical correctness.

**Human evaluation** fails because evaluators lacking dual-domain expertise cannot reliably distinguish correct causal reasoning from superficially plausible domain transfer. Moreover, human raters frequently exhibit a "fluency bias," rewarding models for authoritative technical vocabulary and structured explanations even when the underlying logic is flawed (Casper et al. 2023). This is exacerbated by expert sycophancy, where models generate outputs that align with users' preference for 'expert tone' rather than the ground truth of the target domain (Gudibande et al. 2023). Because RLHF training prioritizes such 'convincing' outputs, the system effectively learns to optimize for perceived expertise rather than causal accuracy.

**Calibration-based detection** fails because expert modules maintain consistent confidence across domain boundaries. Sparse MoE routers are architecturally incentivized to produce low-entropy, peaked gating distributions to satisfy load-balancing and sparsity constraints, which suppresses uncertainty even when inputs are outside the expert's intended domain (Fedus et al. 2022). The design bias toward decisiveness rather than abstention renders standard calibration-based safety filters ineffective, allowing TEE outputs to appear well-calibrated despite being semantically or causally incorrect (Allingham et al. 2023). This problem is exacerbated by a fundamental architectural limitation: in conventional MoE models, expert outputs are fused based on confidence scores that do not reliably reflect true accuracy, resulting in predictions dominated by overconfident experts (D. Zhang et al., 2025). The gating mechanism privileges peaked distributions regardless of whether high confidence reflects genuine domain competence or structural similarity triggering inappropriate pattern completion.

**Adversarial detection** fails because TEE outputs arise from standard model operations. Routing based on learned similarity and generation via pattern completion produce high-confidence outputs even when the domain assignment is incorrect. This phenomenon aligns with 'natural adversarial examples,' (Hendrycks et al. 2021) where ordinary inputs exploit structural biases in the model's weights to trigger confident misclassifications. Because TEE emerges from the model's normal processing of boundary cases, it remains indistinguishable from correct outputs under conventional adversarial checks.

TEE-type hallucinations are particularly problematic for observability because they simultaneously satisfy every conventional reliability signal while being fundamentally incorrect. The four defining characteristics of high confidence, structural plausibility, causal incorrectness, and boundary localization combine to subvert the entire detection stack. Confidence metrics register peaked distributions, coherence checks pass on logical flow, format validators approve correct structure, and human evaluators observe expert-level presentation. The outputs are maximally misleading precisely when they are most wrong: a confident expert voice delivering causally incorrect reasoning in formally impeccable packaging. Because each safety mechanism independently validates different surface properties that TEE preserves, and because no standard method explicitly checks for domain-level causal validity at input-output boundaries, TEE outputs propagate through deployed systems unchecked.

# 7. Mitigation Strategies

Transitive Expert Error in such systems, where domain boundaries are defined by latent space coordinates rather than subjective professional heuristics, can be addressable precisely because the mechanisms are architecturally explicit. This represents a fundamental advantage over human expert systems. Where human pattern recognition and metacognitive failures are observable only through indirect behavioral measures, MoE routers expose selection criteria as weight vectors, expert modules expose activation patterns as probability distributions, and confidence can be measured directly through output entropy. What appears in human cognition as inscrutable intuitive judgments becomes, in MoE architectures, explicit computational operations amenable to direct intervention. We propose three complementary intervention strategies, each targeting specific failure pathways identified in the TEE framework.

## 7.1 Intervention Strategy & Deployment

Deployment context determines intervention priority. *High-stakes applications* in medical, legal, or financial domains should prioritize meta-expert coverage detection and aggressive abstention, accepting reduced coverage to avoid catastrophic false negatives where boundary violations go undetected. *Exploratory applications* like creative tools benefit from boundary-aware calibration over abstention, as users can productively engage with provisional outputs even at domain edges, provided uncertainty is communicated. *Real-time interactive systems* may find multi-expert activation computationally prohibitive and should rely on lightweight meta-experts or post-hoc confidence calibration, accepting higher error rates to maintain latency requirements. These interventions are complementary rather than mutually exclusive: production systems should combine router-level multi-expert activation, specialist-level calibration, and meta-level coverage detection, with thresholds calibrated to deployment requirements.

The routing-coverage distinction determines intervention strategy. Routing-induced failures respond to improved boundary detection: multi-expert activation reveals when specialists disagree, signaling that surface similarity may be deceptive. Coverage-induced failures require abstention mechanisms: when no expert is appropriate, the system must recognize the gap and decline to generate confident outputs. Meta-experts address both pathways by providing architectural capacity to assess whether routing confidence reflects genuine domain match (routing problem) or forced selection from inadequate options (coverage problem).

Each intervention targets specific TEE mechanisms identified in the framework. Multi-expert activation addresses structural similarity bias by forcing comparison between alternative domain models. Boundary-aware calibration addresses authority persistence by training specialists to recognize competence limits. Meta-experts address coverage gaps by providing architectural capacity absent from domain specialists. The interventions are complementary: routing improvements reduce misrouting, calibration reduces overconfidence given correct routing, and meta-experts catch cases where neither applies.

### 7.1.1 Multi-Expert Activation with Disagreement Detection

Standard MoE routing employs a gating network to compute affinity scores between token representations and available experts, selecting top-k based on these scores. Switch Transformer (Fedus et al. 2022) demonstrated the scaling potential of sparse MoE scaling through top-1 routing (achieving constant computational cost per token over trillion-parameter models) but production systems moved to top-2 (Mixtral 8x7B) implicitly recognizing that single-expert confidence can be unreliable, though current implementations use additional experts for averaging rather than boundary detection.

Current routing optimizes exclusively for load balancing and task performance, with no mechanism to distinguish confident routing based on genuine domain match from routing triggered by superficial pattern matching. When causally divergent inputs share surface features, high routing scores may reflect structural similarity bias rather than expert competence. Recent work on dynamic routing (Huang et al. 2024) activates more experts when routing confidence is low, treating this as a signal for additional computation rather than potential boundary violations. Routing-induced TEE emerges precisely here: the router confidently selects based on pattern overlap, the chosen expert confidently generates based on familiar features, and the causal mismatch only becomes apparent through downstream evaluation.

Multi-expert activation with disagreement detection treats divergent outputs as a diagnostic of boundary violations. Research on ensemble disagreement for out-of-distribution detection (Fang et al. 2024; Xu et al. 2024) demonstrates that models trained to different optima exhibit varying loss landscapes on OOD data, with disagreement patterns revealing distributional shifts. When multiple experts with comparable routing scores generate divergent outputs, it indicates that specialists recognize familiar surface patterns but apply inconsistent causal models, signaling structural similarity bias rather than confident consensus on an incorrect output (McCoy et al. 2023). This differs from low-confidence single-expert outputs, which represent standard calibration issues within valid domains (Minderer et al. 2021). In routing-induced TEE, the disagreement itself reveals the structural similarity that misleads the selection mechanism, providing a clear signal for boundary-aware intervention (Fang et al. 2024).

Disagreement is diagnostically meaningful only when experts are invoked with comparable routing confidence; divergence under uniformly low or highly asymmetric routing scores reflects epistemic uncertainty or routing noise rather than boundary failure. Implementation augments standard routing losses (task performance plus load balancing) with boundary detection terms: `L_boundary` penalizes low-entropy routing when inputs are equidistant from multiple expert centroids, and `L_coverage` penalizes confident routing when all affinity scores fall below threshold $\tau$, extending optimization to include explicit boundary recognition.

Training instabilities can further degrade routing reliability. Zheng et al. (2025) show that expert activations can shift dramatically between policy updates even for identical inputs, indicating that routing confidence may reflect training dynamics rather than genuine domain expertise. Stable routing is prerequisite for disagreement-based boundary detection.

7.1.2 **Boundary-Aware Calibration**

Even with correct routing, experts may generate confidently incorrect outputs when inputs approach domain boundaries or when surface features trigger inappropriate pattern completion. Modern neural networks are poorly calibrated, producing overconfident predictions despite high accuracy, with depth, width, weight decay, and Batch Normalization identified as factors influencing miscalibration (Guo et al. 2017). This overconfidence becomes particularly problematic at domain boundaries where experts encounter inputs that share surface features with their training distribution but require different causal models.

Calibration research focuses primarily on post-training correction methods. Temperature scaling, a post-processing technique that divides logits by a learned scalar parameter to soften output distributions, has proven remarkably effective at restoring calibration for in-distribution predictions (Guo et al. 2017) but only works when the test distribution matches the training distribution, and is not guaranteed to produce calibrated probabilities on out-of-distribution samples. More recent work has explored adaptive temperature scaling methods that compute different temperature factors for each prediction based on entropy (Balanya et al. 2024), recognizing that models exhibit varying levels of overconfidence depending on input characteristics rather than uniform miscalibration across all predictions.

Training-time interventions provide an alternative approach. Confidence regularization (Pereyra et al. 2017) penalizes overconfident predictions during training through entropy-based penalties that discourage peaked distributions on ambiguous inputs. However, these approaches treat calibration as a general robustness problem rather than specifically targeting expert competence boundaries in specialized models. Research on fine-tuning and out-of-distribution generalization reveals a fundamental tension: fine-tuning can achieve worse accuracy than linear probing out-of-distribution when pretrained features are good and the distribution shift is large, obtaining on average 2% higher accuracy in-distribution but 7% lower accuracy out-of-distribution (Kumar et al. 2022). This suggests that standard fine-tuning procedures may actually degrade the boundary recognition capabilities we seek to preserve.

Boundary-aware calibration for MoE experts extends these techniques by training on synthesized boundary cases where surface similarity masks causal divergence. For a computer science expert in an MoE system, this means learning to signal uncertainty not just on random noise or clearly foreign inputs, but specifically on boundary cases like queries about biological neural networks, social networks, or fishing nets that share vocabulary while requiring different reasoning frameworks. Experts learn to generate flattened probability distributions on such inputs, signaling epistemic uncertainty through higher entropy rather than maintaining peaked distributions characteristic of confident within-domain responses Malinin & Gales (2018). This requires training data that includes boundary cases explicitly: inputs designed to trigger structural similarity bias through shared vocabulary with divergent causality, teaching experts to recognize when their pattern-matching machinery encounters familiar features in inappropriate contexts.

### 7.1.3 Meta-Experts for Boundary Recognition

A meta-expert is a specialized module trained to detect domain boundaries and coverage gaps rather than to solve domain-specific problems. This addresses coverage-induced TEE by providing an architectural mechanism for recognizing when no available expert has appropriate training. Current MoE systems optimize individual experts for in-domain perplexity and routers for affinity-based selection; consequently, every component assumes domain appropriateness while no component verifies it. A meta-expert fills this structural gap by assessing whether routing confidence reflects genuine competence or a forced selection from inadequate options.

This concept synthesizes two research directions. First, meta-learning for domain generalization (Li et al. 2019) utilizes episodic training to handle out-of-distribution data by simulating domain shifts. Second, AI safety research on scalable oversight (Bowman et al. 2022) utilizes auxiliary models to monitor the performance of complex systems. However, the meta-expert specifically addresses the MoE architectural deficit: it provides the explicit boundary detection that neither generative experts nor similarity-based routers are optimized to perform.

The meta-expert receives either the original query embedding or the concatenation of candidate expert outputs to predict whether the input falls within the system's collective competence domain. Unlike domain experts, its training data consists of examples explicitly labeled for domain membership, boundary cases, and coverage gaps that data domain specialists never encounter. This functions as a form of architectural regularization; just as weight penalties prevent overfitting to training points, the meta-expert's boundary-focused training prevents the system from overconfidently extrapolating beyond its validated expertise.

Implementation utilizes lightweight architectures, such as a small transformer or MLP operating over embeddings, since the task is classification rather than generation. The meta-expert outputs reliability assessments that can trigger system-level responses: explicit abstention, requests for context, or fallback to generalist processing. When integrated with multi-expert disagreement detection (Section 7.1.1), it provides a redundant safety signal, combining internal architectural divergence with external oversight to identify boundary violations with high confidence.

### 7.1.4 Detecting Coverage-Induced Failures

Coverage-induced failures require system-level detection when no available expert has appropriate training for an input, combining multiple signals, not relying on any single metric.

Expert-level OOD scoring measures the distance of inputs from each expert's training distribution in latent space (Hendrycks & Gimpel 2017). High distances across all experts indicate coverage gaps. However, distance metrics alone are insufficient because they cannot distinguish genuine coverage gaps from boundary cases where an appropriate expert exists but similarity scores are ambiguous (Liang et al. 2018).

Inter-expert disagreement provides complementary information (Malinin & Gales 2018). When multiple experts produce divergent outputs with comparable confidence, the system infers that no single expert has clear competence. This differs from confident consensus around a wrong model, as in routing-induced TEE when all experts make the same category error. High dissent ('predictive variance') combined with high OOD scores provides stronger evidence for coverage gaps than ambiguous routing or confident error (Lakshminarayanan et al. 2017).

Deployment responses to detected coverage gaps can include explicit uncertainty statements in outputs, caveats about domain limitations, or refusal to generate confident responses (Stutz et al. 2022). The key is making coverage gaps visible to users rather than masking them with confident but unreliable outputs characteristic of undetected TEE. Users can then decide whether to reformulate queries, provide additional context, or seek alternative information sources.

### 7.1.5 Architectural Stability for Multi-Stream Expert Systems

The effectiveness of these interventions depends on the underlying stability and training of the MoE system: without well-calibrated experts and robust embeddings, multi-expert activation, meta-experts, and boundary-aware calibration cannot function reliably. The architectural stability question addresses a foundational prerequisite for TEE mitigation: ensuring that expert systems train reliably at scale. Training instabilities that corrupt expert development can produce poorly defined domain boundaries and unreliable confidence calibration, undermining boundary-aware calibration (7.1.2), multi-expert disagreement detection (7.1.1), and meta-expert oversight (7.1.3). While not designed specifically to address boundary violations, stable training determines whether experts develop the coherent specializations and appropriate confidence calibration that other mitigations depend upon.

Recent approaches exemplified by Hyper-Connections have extended the residual connection paradigm, expanding the residual stream width and diversifying connectivity patterns, yielding substantial performance gains but compromise the identity mapping property intrinsic to residual connections, causing severe training instability and restricted scalability (Zhu et al. 2024). This instability manifests as loss spikes and gradient explosions in training, particularly problematic for MoE architectures where many experts must develop distinct specializations simultaneously.

Manifold-Constrained Hyper-Connections (mHC) projects residual connection matrices onto a doubly stochastic manifold using the Sinkhorn-Knopp algorithm, restoring identity mapping and stabilizing signal propagation through geometric constraints that ensure convex combination-based feature flow (Xie et al. 2025). The doubly stochastic constraint ensures both row and column sums equal 1, making operations convex combinations of input features that conserve feature mean and strictly regularize signal norm, effectively mitigating vanishing or exploding signals. The closure property of doubly stochastic matrices under multiplication ensures that composite mappings across arbitrary depths maintain this conservation property, providing stable training as models scale to hundreds of billions of parameters.

Empirical experiments demonstrate that mHC trains smoothly while unconstrained Hyper-Connections exhibit unexpected loss surges around 12,000 steps correlated with gradient norm instability, with the framework achieving this stability while introducing only approximately 6.7% training overhead. While mHC was developed to enable stable training at scale rather than to mitigate cross-domain errors specifically, stable training is necessary infrastructure for other TEE mitigations to function effectively. When gradient explosions or signal collapse occur, experts may fail to learn appropriate uncertainty at domain edges or develop sufficiently distinct specializations for disagreement to signal boundary violations.

The distinction between architectural stability and architectural oversight (7.1.3) is crucial: stability mechanisms prevent training failures that would produce poorly calibrated experts, while oversight mechanisms detect and respond to boundary violations during inference with well-trained experts. Both are necessary components of comprehensive TEE mitigation.

### 7.1.6 Training Enhancements

The mitigation strategies that require training data and objectives that make domain boundaries explicit rather than leaving them implicit in statistical patterns.

Contrastive domain learning trains embeddings to separate superficially similar but causally divergent cases (Chen et al. 2020; Khosla et al. 2020). Pairs like 'neural networks in neuroscience' versus 'neural networks in machine learning' are explicitly contrasted during training, teaching the embedding space to encode causal domain distinctions that naive statistical co-occurrence obscures, to better reflect domain boundaries than token distributions alone.

Synthetic boundary augmentation introduces training examples where shared vocabulary masks different causal mechanisms (Hendrycks et al. 2020). These examples teach both routers and experts to recognize surface similarity as potentially deceptive rather than automatically triggering confident domain application. The augmentation can be systematic, using domain taxonomies to identify boundary regions, or adversarial, actively searching for cases where current routing produces linearly confident but incorrect outputs (Goodfellow et al. 2015).

## 7.2 Measurable Success Criteria

Multi-expert activation generates testable predictions: (1) disagreement rate should correlate with human expert identification of boundary cases, (2) precision on committed predictions should substantially exceed single-expert routing on labeled boundary datasets, (3) coverage should decrease as abstention increases on ambiguous inputs.

Boundary-aware calibration should produce measurable correlation between expert output entropy and domain inappropriateness. An ideal system shows monotonicity: as inputs move from expert training domains, entropy increases proportionally. This is directly measurable by computing entropy-distance correlations across boundary cases with known domain labels.

Meta-expert boundary detection should outperform confidence thresholding on individual experts by substantial margins in precision-recall, leveraging specialized boundary detection versus naive uncertainty measures.

Critically, these metrics are observable during inference without ground truth labels. Disagreement rates, entropy distributions, and meta-expert confidence provide real-time TEE signals, enabling continuous monitoring and adaptive intervention thresholds. Systems can track these metrics over time, identifying domains where boundary violations increase and adjusting routing or calibration accordingly.

The interventions enable production A/B testing. Deployments with and without multi-expert activation can be compared on user engagement, error rates, and task success. Boundary-aware calibration can be evaluated by whether users prefer appropriately uncertain responses over confident errors. Meta-expert effectiveness shows in precision of flagged cases that evaluators identify as boundary violations.

## 7.3 Toward Tractable Expert Systems

These strategies increase computational costs via additional inference, boundary augmentation, calibration objectives, and meta-expert architecture, but provide concrete pathways toward MoE systems that detect and mitigate TEE rather than confidently executing in wrong domains. Each intervention targets specific failure mechanisms identified here, making the problem tractable through architectural design rather than assuming scale resolves boundary violations.

Scale amplifies expert capability within domains while providing no mechanism for boundary recognition. More capable experts produce more convincing wrong answers at boundaries, with higher technical sophistication, lower perplexity, and yet: causally incorrect. These interventions convert undetectable human-expert over-exuberance into measurable system behavior with explicit intervention points. Where human TEE requires institutional scaffolding like epistemic auditors and qualification procedures, MoE TEE can be addressed through architecture, making reliable expert systems tractable for the first time.

# 8. Future Directions

## 8.1 The Hard Problem: Discrete Specialization in Continuous Space

TEE in MoE systems reveals a fundamental tension: we want discrete expert specialization (clear domains with bounded competence) but operate in continuous representation spaces (embeddings, activation patterns). Every input has some similarity to every expert's training distribution. The salient question is whether that similarity reflects genuine domain overlap or superficial pattern matching. The routing mechanism has access only to surface features (tokens, embeddings, semantic similarity) while domain appropriateness depends on causal structure not represented in the routing space.

This isn't just an MoE problem. Any sufficiently complex model exhibits implicit specialization. Dense models face the same challenge but implicitly: structural similarity activates wrong circuits without explicit routing decisions. MoE makes the problem visible and potentially addressable.

The gating problem is harder than the expert problem. Even with perfectly calibrated specialists, structural similarity makes accurate routing fundamentally difficult. This challenges the assumption that expert mixture improves over dense models: mixture provides benefits within clear domains but may worsen performance at boundaries where routing errors trigger confident wrong outputs.

Standard scaling laws assume that more parameters and broader training improve performance monotonically. TEE suggests this may not hold at domain boundaries. As models become more capable within domains, authority persistence strengthens: the expert that is genuinely sophisticated in computer science will be more confidently wrong about fishing nets than a superficial model. The metacognitive halo effect intensifies with capability. Scaling up expert quality may actually worsen boundary failures unless accompanied by improved metacognitive calibration.

This has implications for training strategies. Current approaches optimize within-domain performance, which increases expert capability and confidence. But without explicit boundary training, this exacerbates TEE. Future work must explore training objectives that trade some within-domain performance for improved boundary recognition.

## 8.2 Questions for Further Investigation

**Architectural questions**: What granularity of expert specialization minimizes TEE? Fine-grained experts reduce domain size but increase boundary surface area. Coarse-grained experts reduce boundaries but require individual experts to span more domain diversity. Is there an optimal granularity or does it depend on domain structure?

**Representation questions**: Can causal structure be represented explicitly in ways accessible to routing mechanisms? Current routers operate on statistical features. What would it take to represent domain-specific causal architectures such that routers could assess mechanistic fit rather than just pattern similarity?

**Detection questions**: Can TEE signatures be detected post-hoc in deployed systems? The confidence-accuracy dissociation and domain-inappropriate content patterns suggest automated detection might be possible. This could enable safety monitoring without requiring architectural changes to existing models.

**Limits questions**: Is there a fundamental limit to domain boundary detection in high-dimensional continuous spaces? Or can sufficiently sophisticated architectures learn to reliably distinguish surface similarity from causal fit? This question bears on long-term AI safety: if reliable boundary detection is impossible, then any sufficiently complex system will exhibit TEE-type failures as capability increases.

**Generalization questions**: Do dense models (transformers without explicit routing) exhibit analogous failures through implicit circuit specialization? If so, TEE is a fundamental challenge for any sufficiently capable system, not just MoE. How do other architectures such as retrieval-augmented generation, tool-using agents, constitutional AI face similar boundary problems?

The most critical question is whether TEE intensifies or diminishes as expert modules become more capable. If more sophisticated experts develop better uncertainty estimation, TEE may be a transitional problem that diminishes with scale. But if expert capability within domains increases output confidence without corresponding improvements in boundary recognition, TEE becomes more severe: more sophisticated experts produce more convincing wrong answers at boundaries, like human experts. This scaling relationship depends on whether training objectives reward within-domain performance alone or also penalize confident execution on out-of-domain inputs.

## 8.3 Cross-Architecture Applications

The TEE framework applies directly to emerging AI system architectures:

**Multi-model systems** (e.g., routing between GPT-4, Claude, Gemini based on task type): Do orchestration layers exhibit the same routing-induced and coverage-induced failures? Can meta-models detect when surface task descriptions mask domain boundaries? The interventions translate directly: multi-model activation with disagreement detection, confidence calibration to model training domains, explicit abstention when no model is appropriate.

**Tool-using agents** (e.g., AutoGPT, ReAct frameworks selecting from API catalogs): Tools have sharp competence boundaries, making coverage failures particularly acute. Does tool selection based on function descriptions create systematic misapplication when terminology is shared? The meta-expert approach translates: a specialized module could assess tool-task fit beyond surface feature matching.

**Compound AI systems** (e.g., systems routing between retrieval, generation, verification modules): Do these systems confidently route financial queries to physics-trained retrievers when terminology overlaps? The boundary-aware calibration approach applies: training components to signal uncertainty when inputs diverge from training domains.

**RAG with specialized corpora**: When query routing selects knowledge sources, does shared vocabulary create routing-induced failures? The multi-source activation with disagreement detection directly applies.

Each architecture instantiates the same structural problem—*routing based on surface features when domain appropriateness depends on causal structure*—making TEE a fundamental challenge for all specialized AI systems, not merely an MoE phenomenon.

# 9. Conclusion

When MoE routers send fishing tokens to computer science experts based on shared vocabulary, they make the same error as physicists applying deterministic models to reflexive financial markets: pattern recognition optimized for one causal structure executes confidently on inputs where surface similarity masks fundamental divergence. Transitive Expert Error operates through structural similarity driving inappropriate selection, authority persisting across competence boundaries, with delayed feedback preventing recalibration.

MoE architectures make this problem tractable by exposing what remains hidden in human cognition. Routers weight experts based on token distributions. Expert modules output peaked probability distributions on out-of-domain inputs. RLHF correlates with coherence independent of domain appropriateness. The proposed interventions (multi-expert routing with disagreement detection, boundary-aware calibration, meta-experts for boundary recognition) are implementable because MoE architectures make the gating problem explicit.

Current architectures optimize for computational efficiency, load balancing, and within-domain accuracy without examining whether routing reflects causal domain fit or whether confidence calibrates to competence boundaries. This structural indifference drives TEE.

TEE in artificial systems has observable signatures: routing patterns based on lexical overlap, confidence-accuracy dissociations on out-of-domain inputs, domain-inappropriate content detectable through automated analysis. Unlike human expert judgment where failures are identified retrospectively, MoE systems enable real-time monitoring.

The failure mode of confident, coherent, structurally plausible but causally incorrect outputs at domain boundaries is not an edge case. Similar to Oliver Hart's demonstration of contract residuals (which are inherently ambiguous) any sufficiently complex application will encounter queries spanning domains or falling into coverage gaps. Scaling does not resolve TEE; better training data does not resolve TEE absent explicit boundary-aware objectives. More capable experts intensify authority persistence, amplifying the failure mode at boundaries.

The proposed interventions are implementable in current architectures using straightforward extensions of current practices: boundary case augmentation, coverage gap labeling, confidence penalties for low-similarity routing. Multi-expert routing with disagreement flagging identifies coverage gaps reliably enough to trigger explicit abstention or generalist fallback.

The broader implication extends to any system routing inputs between specialized processors: subnetworks within MoE, distinct models in inference pipelines, services in compound AI systems. All face the same gating problem when learned representations privilege surface patterns over causal domain fit. Structural similarity in high-dimensional representations is not going away, but it does not have to produce systematic hallucination at boundaries. The mechanisms are known, the signatures are detectable, and the interventions are implementable.

Systems being built today will either treat domain boundaries as first-class architectural concerns or they will reliably produce confident, causally incorrect outputs whenever structural similarity is equivocal.